\documentclass[letterpaper, 10 pt, conference]{ieeeconf}  

\IEEEoverridecommandlockouts                              

\usepackage{graphics} 
\usepackage{epsfig} 
\usepackage{mathptmx} 
\usepackage{times} 
\usepackage{amsmath} 
\usepackage{amssymb}  
\usepackage{adjustbox}
\usepackage{mathtools}
\usepackage{hyperref}
\usepackage{booktabs}
\usepackage{array}
\usepackage{indentfirst}
\usepackage{cite}
\usepackage{array}
\usepackage{booktabs}
\usepackage{multirow}
\usepackage{graphicx} 
\usepackage{bm}

\title{\LARGE \bf
BEAM: Brainwave Empathy Assessment Model for Early Childhood
}

\author{
    Chen Xie$^{1}$, Gaofeng Wu$^{1}$, Kaidong Wang$^{1}$, Zihao Zhu$^{1}$, Xiaoshu Luo$^{1}$, Yan Liang$^{1}$,\\ Feiyu Quan$^{1}$, Ruoxi Wu$^{2}$, Xianghui Huang$^{3}$, Han Zhang$^{1,4,5}$
    \thanks{
        This work was supported by the STI 2030-Major Projects (No. 2022ZD0209000), the Shanghai Pilot Program for Basic Research - Chinese Academy of Sciences, Shanghai Branch (No. JCYJ-SHFY-2022-014), the Key Program of Xiamen Medical and Health (Grant No. 3502Z20234013), and the Grand Strategic Project of the National Social Science Fund of China (No. 23\&ZD319). The computation in this work was also supported by the HPC Platform of ShanghaiTech University.
    }
    \thanks{
        $^{1}$School of Biomedical Engineering, ShanghaiTech University, Shanghai, China. 
        {\tt\small xiechen2023@shanghaitech.edu.cn}
    }
    \thanks{
        $^{2}$Ministry of Brain Functional Genomics (MOE\&STCSM), School of Psychology and Cognitive Science, East China Normal University, Shanghai, China.
    }
    \thanks{
        $^{3}$Children's Hospital of Fudan University (Xiamen Branch), Xiamen Children's Hospital, Fujian Key Laboratory of Neonatal Diseases, Xiamen, Fujian, China.
    }
    \thanks{
        $^{4}$State Key Laboratory of Advanced Medical Materials and Devices, ShanghaiTech University, Shanghai 201210, China. 
        {\tt\small zhanghan2@shanghaitech.edu.cn}
    }
    \thanks{
        $^{5}$Shanghai Clinical Research and Trial Center, Shanghai 201210, China. }
}

\begin{document}

\maketitle
\thispagestyle{empty}
\pagestyle{empty}

\begin{abstract}

Empathy in young children is crucial for their social and emotional development, yet predicting it remains challenging. Traditional methods often only rely on self-reports or observer-based labeling, which are susceptible to bias and fail to objectively capture the process of empathy formation. EEG offers an objective alternative; however, current approaches primarily extract static patterns, neglecting temporal dynamics. To overcome these limitations, we propose a novel deep learning framework, the Brainwave Empathy Assessment Model (BEAM), to predict empathy levels in children aged 4 to 6 years. BEAM leverages multi-view EEG signals to capture both cognitive and emotional dimensions of empathy. The framework comprises three key components: 1) a LaBraM-based encoder for effective spatio-temporal feature extraction, 2) a feature fusion module to integrate complementary information from multi-view signals, and 3) a contrastive learning module to enhance class separation. Validated on the CBCP dataset, BEAM outperforms state-of-the-art methods across multiple metrics, demonstrating its potential for objective empathy assessment and providing a preliminary insight into early interventions in children’s prosocial development.

\noindent\hspace*{1em}\textit{\textbf{Index Terms}}---EEG, deep learning, empathy, children.

\end{abstract}

\section{INTRODUCTION}

Empathy plays a crucial role in facilitating the transmission of emotional states \cite{c1}, underpinning caregiving and prosocial behaviors, whereas its absence relates to severe social-emotional dysfunctions, such as antisocial personality disorder \cite{c2}. Empathy can be divided into two key components: cognitive empathy, also known as Theory of Mind (ToM), which refers to the ability to understand and adopt another’s perspective, and emotional empathy (EM), which involves the ability to share the emotional states of others in both valence and intensity \cite{c3}. While empathy-related abilities begin to emerge in infancy, their full development relies on continuing support through early emotional interactions with caregivers\cite{c4}. In general, assessing empathy in children is critical for understanding key stages of social-emotional development and provides a scientific basis for early interventions to foster healthy emotional and social behaviors.

Recent cognitive neuroscience research has increasingly emphasized the study of empathy development in children\cite{c5}. However, accurately assessing and predicting empathy levels in children remains a significant challenge. Existing assessments are primarily based on subjective methods, such as self-reports\cite{c6}, parent surveys\cite{c5}, and manual annotations\cite{c7}, which lack objectivity and real-time monitoring capabilities. Moreover, current research often treats empathy as a single-dimensional construct, neglecting comprehensive modeling of cognitive and emotional components. While EEG has been employed in empathy prediction studies, these efforts are predominantly focused on adults and are limited to extracting static asymmetry features\cite{c8}, thereby overlooking critical dynamic information and spatial patterns.

To address these challenges, we propose a robust deep learning framework that uses EEG signals to objectively and efficiently assess children’s empathy levels. Based on evidence that video stimuli evoke ToM and EM in children aged 3–12 years \cite{c9}, and that EEG signals reflect neural correlates of empathetic prosocial behaviors \cite{c10}, our approach models both dimensions of empathy. EEG signals from ToM and EM events are processed through Large Brain Model (LaBraM) encoder to extract spatial and dynamic temporal representations at global and local levels. ToM and EM, as complementary components of empathy, are further integrated using a feature fusion module that preserves shared features while capturing component-specific characteristics. Additionally, a contrastive learning module is applied to improve cross-subject consistency and enhance prediction performance.

The key contributions of this study are summarized as follows:
\begin{enumerate}

    \item \textbf{Children Empathy Prediction:} To our best knowledge, we unprecedentedly leverage EEG signals to predict children's willingness to help, a novel classification approach provides new insight into empathy development and may support early interventions in evaluating children's prosocial behaviors.

    \item \textbf{Multi-view Framework with Contrastive Learning:} We propose a multi-view framework that integrates ToM and EM dimensions from EEG signals through feature fusion for a robust empathy representation. The encoder captures temporal dynamics and preserves spatial information for high-quality feature extraction. Additionally, contrastive learning further enhances performance by reducing cross-subject variability.

    \item \textbf{Comprehensive validation:} We comprehensively validate our proposed method, BEAM, on the ongoing Chinese Baby Connectome Project (CBCP) and achieve superior performance compared to other state-of-the-art (SOTA) methods, demonstrating its effectiveness in empathy prediction for young children.
\end{enumerate}

\section{Method}

\subsection{Datasets}
We validate the proposed method on the CBCP dataset, which includes EEG data from 57 typically developing children (aged \( 4.91 \pm 1.07 \) years). Each participant undergoes a 6-minute session, during which EEG signals are recorded using 32 channels at a 1000 Hz sampling rate, following the international 10-20 Brain Products (BP) system. 

During the session, participants watch Pixar’s \textit{Partly Cloudy}\footnote{Available at \href{https://www.pixar.com/partly-cloudy}{https://www.pixar.com/partly-cloudy}} (2009), a 5-minute animated film. After watching the film, participants are asked to complete a post-test guided by pictures and audio, assessing their empathy levels by their willingness to help in the negative scenario. The willingness-to-help scores range from 1 to 4. As outlined in \cite{c8}, a median split was used to divide self-reported empathy scores into high- and low-empathy groups. The study has received approval from the Research Ethics Committees of the institutes of the corresponding author. This trial is registered with ClinicalTrials.gov (NCT05040542).

\subsection{Preprocessing}
 
Data preprocessing is conducted with EEGLAB in Matlab2022b \cite{c11}, including bandpass filtering (0.1–75 Hz) to retain relevant frequencies, downsampling to 200 Hz to reduce data dimensionality, independent component analysis (ICA) to remove artifacts, and re-referencing to common average to enhance channel consistency. Following the video segmentation method in \cite{c9}, the recordings are divided into 6 clips for ToM Events and 8 clips for EM Events.

The willingness-to-help scores from the post-test self-assessments exhibited a slight class imbalance. To mitigate this issue, we employ data augmentation techniques. Specifically, we apply short-time Fourier transform (STFT) to EEG data, then perturb the frequency domain amplitudes with random Gaussian noise (mean = 0, std = 0.001) . The signals are then reconstructed using inverse STFT\cite{c12}.

\subsection{Framework}
\subsubsection{Input}
We aim to predict children’s willingness to help by analyzing EEG signals recorded from the same subject with both ToM and EM events. ToM and EM represent cognitive and emotional empathy, the two fundamental components of empathy.

To address inconsistent event lengths and reduce computational costs, we segment each event into smaller samples. Following the approach in \cite{c13}, for multi-channel EEG signals \( \bm{E} \in \mathbb{R}^{C \times T}\),
where $C$ is the number of electrodes and $T$ is the total timesteps, segmentation was performed using a window length of $W = 4s$ and a stride of $S = 1s$. This method divides \( \bm{E} \) into \( \lfloor (T-W)/S \rfloor + 1 \)
samples, each represented as \( \bm{X}  \in \mathbb{R}^{C \times W} \). Finally, each subject generates 65 ToM and 43 EM samples.

\subsubsection{Network architecture}
\begin{figure*}[htbp]
    \centering
    \includegraphics[width=1\textwidth]{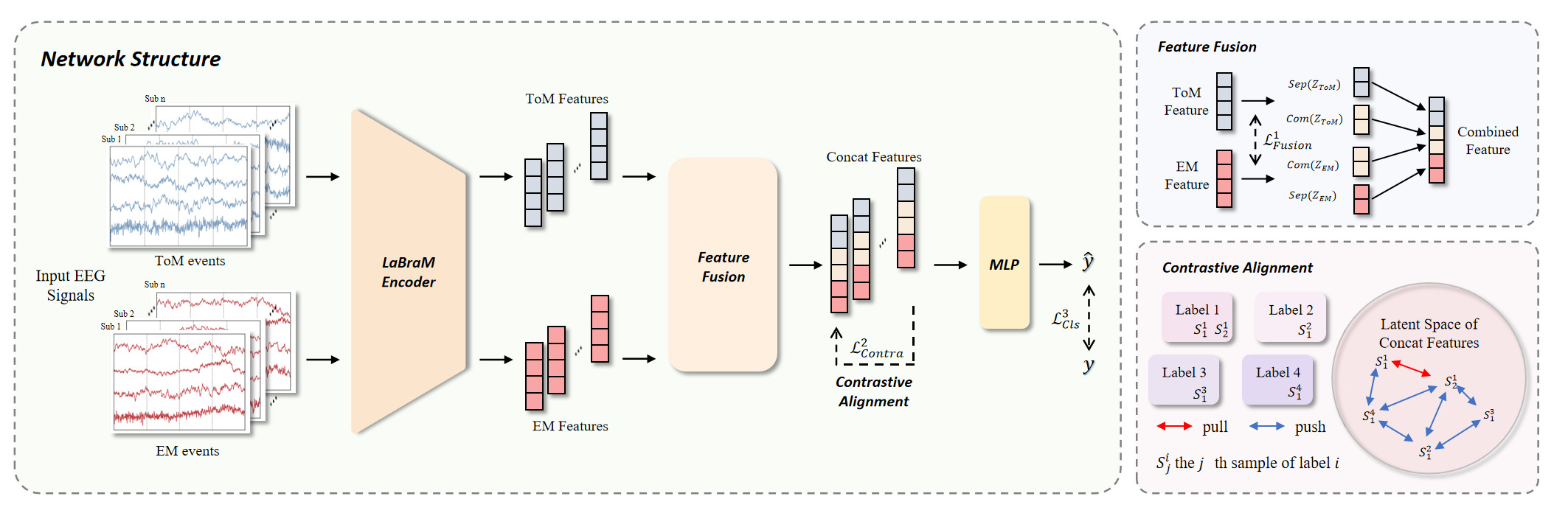} 
    \caption{Overall architecture of the proposed method. }
    \label{figurelabel}
\end{figure*}

\paragraph{Encoder}
Building upon the LaBraM framework in \cite{c13}, our method is specifically tailored to predict children’s empathy levels by capturing and integrating multi-view signals. The overall architecture is illustrated in Fig. \ref{figurelabel}.

 LaBraM \cite{c13} is a large-scale EEG model based on a neural Transformer, pre-trained in an unsupervised manner on over 2,500 hours of diverse EEG data, enabling a comprehensive understanding of universal EEG signals. Input EEG signals $X \in \mathbb{R}^{C \times W}$are segmented into fixed-length windows to extract temporal features. Spatial-temporal embeddings are applied to each patch, which is subsequently processed by the Transformer encoder, allowing integration of both global and local information. The encoder outputs two distinct representations, $Z_{\text{ToM}}$ and $Z_{\text{EM}}$, which correspond to the cognitive and emotional empathy dimensions, respectively.

\paragraph{Feature Fusion}
We perform feature fusion after extracting features from each view to better integrate complementary information between ToM and EM. Specifically, we decomposed the view-specific latent feature \( \mathbf{Z}_{\text{ToM}} \) and \( \mathbf{Z}_{\text{EM}} \)  into two parts: a shared component \( \mathrm{Com}(\mathbf{Z}_n) \) and a specific component \( \mathrm{Sep}(\mathbf{Z}_n) \), where \( n \in \{ToM, EM\} \). \( \mathrm{Com}(\mathbf{Z}_n) \). \( \mathrm{Com}(\mathbf{Z}_n) \) represents the shared information among modalities, while \( \mathrm{Sep}(\mathbf{Z}_n) \) represents the complementary information that differentiates one view from another. The fusion process follows the basic requirements:

\begin{enumerate}
    \item \( \mathbf{Z}_n = [ \mathrm{Com}(\mathbf{Z}_n), \mathrm{Sep}(\mathbf{Z}_n) ] \).
    \item \( \mathrm{Com}( \mathbf{Z}_{\text{ToM}} ) \) and \( \mathrm{Com}( \mathbf{Z}_{\text{EM}} ) \) are as similar as possible.
    \item \( \mathrm{Sep}( \mathbf{Z}_{\text{ToM}} ) \) and \( \mathrm{Sep}( \mathbf{Z}_{\text{EM}} ) \) are as distinct as possible.
\end{enumerate}

Accordingly, the fusion loss \( \mathcal{L}_{\mathrm{Fusion}} \) is defined as:
\begin{equation} 
\mathcal{L}_{\mathrm{Fusion}} = \frac{\left| \mathrm{Sim}_{\mathrm{Sep}} \right|}{\mathrm{Sim}_{\mathrm{Com}} + 1 + \epsilon}
\end{equation}
where
\begin{equation}
\text{Sim}_{\text{Com}} = \frac{\text{Com}(\mathbf{Z}_{\text{ToM}}) \cdot \text{Com}(\mathbf{Z}_{\text{EM}})}{\|\text{Com}(\mathbf{Z}_{\text{ToM}})\| \|\text{Com}(\mathbf{Z}_{\text{EM}})\|} 
\end{equation}

\begin{equation}
\text{Sim}_{\text{Sep}} = \frac{\text{Sep}(\mathbf{Z}_{\text{ToM}}) \cdot \text{Sep}(\mathbf{Z}_{\text{EM}})}{\|\text{Sep}(\mathbf{Z}_{\text{ToM}})\| \|\text{Sep}(\mathbf{Z}_{\text{EM}})\|}
\end{equation}

After feature fusion, the fused representation is formed as:
\begin{equation} 
\mathbf{Z}_{ToM, EM} = \left( \mathrm{Sep}( \mathbf{Z}_{\text{ToM}} ), \mathrm{Common}, \mathrm{Sep}( \mathbf{Z}_{\text{EM}} ) \right)
\end{equation}
where \( \mathrm{Common} = 0.5 \left( \mathrm{Com}( \mathbf{Z}_{\text{ToM}} ) + \mathrm{Com}( \mathbf{Z}_{\text{EM}} ) \right) \).

\paragraph{Contrast Learning}
To effectively capture discriminative features, we applied contrastive learning with InfoNCE loss\cite{c16} to maximize inter-class separability while minimizing intra-class variance. Given a batch of feature representations \( \mathbf{Z} = [z_1, z_2, \dots, z_B] \, \text{with batch size } B \). The InfoNCE loss is defined as:

\begin{equation}
\mathbf{Z}_{\text{norm}} = \frac{\mathbf{Z}}{\|\mathbf{Z}\|_2}
\end{equation}

\begin{equation}
\mathcal{L}_{\text{Contra}} = -\frac{1}{B} \sum_{i=1}^B \log \left( 
\frac{\exp\left(\frac{\mathbf{z}_i \cdot \mathbf{z}_{i+}}{\tau}\right)}{\sum_{j=1}^B \exp\left(\frac{\mathbf{z}_i \cdot \mathbf{z}_j}{\tau}\right)} 
\right)
\end{equation}

where:
\begin{enumerate}
    \item $\mathbf{z}_i$ is the encoded vector of the query sample $i$.
    \item  $\mathbf{z}_{i+}$ is the positive sample corresponding to the query sample $i$.
    \item \( \tau \) is the temperature parameter controlling softmax sharpness.
    
\end{enumerate}

\subsection{Performance Evaluation}
To avoid information leakage, we divided the dataset at the subject level. Specifically, the dataset was divided into training (70\%), validation (20\%), and testing (10\%). All experiments were repeated five times with different random seeds to ensure the robustness and reliability of the results.

The evaluation metrics included accuracy, sensitivity, and specificity.




\section{Result}

\subsection{Comparison with SOTA Models}

We compare our model with SOTA methods, including BIOT, ST-Transformer, and SVM-asymmetry. Table \ref{table_performance_comparison} shows that the proposed method outperforms other methods across all metrics. This indicates a performance breakthrough for BEAM (Proposed Method). The smaller standard deviation suggests that BEAM demonstrates good stability across experiments, ensuring reliability.

\begin{table}[h]
\caption{Performance Comparison}
\scriptsize
\centering
\label{table_performance_comparison}
\begin{adjustbox}{max width=\textwidth}
\setlength{\tabcolsep}{3pt}
\begin{tabular}{|c|c|c|c|}
\hline
\textbf{Method} & \textbf{Accuracy}~$\uparrow$ & \textbf{Specificity}~$\uparrow$ & \textbf{Sensitivity}~$\uparrow$ \\
\hline
ST-Transformer \cite{c4} & 0.512$\pm$0.021 & 0.511$\pm$0.022 & 0.512$\pm$0.022 \\
\hline
SVM-asymmetry \cite{c8} & 0.538$\pm$0.001 & 0.533$\pm$0.002 & 0.536$\pm$0.002 \\
\hline
BIOT \cite{c14} & 0.564$\pm$0.012 & 0.571$\pm$0.018 & 0.560$\pm$0.016 \\
\hline
\textbf{Proposed Method} & \textbf{0.647$\pm$0.008} & \textbf{0.651$\pm$0.009} & \textbf{0.646$\pm$0.009} \\
\hline
\end{tabular}
\end{adjustbox}
\end{table}

\subsection{Ablation Experiments}
To evaluate the effectiveness of network modules and multi-view feature components, we conduct two sets of ablation experiments:

\subsubsection{Effectiveness of Empathy Component}

Table \ref{table_empathy_components} presents ablation results for empathy components. The EM component achieves an accuracy of 0.588$\pm$0.002, while ToM achieves 0.614$\pm$0.007, highlighting ToM's greater impact. Combining EM and ToM with contrastive learning module improves accuracy by 2.5\% compared to ToM alone, and even the simple concatenation of EM and ToM outperforms single-component inputs without advanced fusion techniques. Contrastive learning boosts most metrics, with EM showing slightly more improvement, indicating its positive impact. The lower variance in EM suggests stable performance, but contrastive learning yields limited improvement, reflecting the constraints of EM in this model.

\begin{table}[h]
\scriptsize
\centering
\caption{Ablation Experiments of Empathy Components}
\label{table_empathy_components}
\begin{adjustbox}{max width=\textwidth}
\setlength{\tabcolsep}{3pt}
\begin{tabular}{|c|c|c|c|c|}
\hline
\multicolumn{2}{|c|}{\textbf{Method (contrast)}} & \textbf{Accuracy$\uparrow$} & \textbf{Specificity$\uparrow$} & \textbf{Sensitivity$\uparrow$} \\
\hline
\multirow{2}{*}{\textbf{EM}}     & ×                                   & 0.588$\pm$0.002              & 0.592$\pm$0.002              & 0.577$\pm$0.002              \\
                                 & \checkmark                          & 0.592$\pm$0.003              & 0.588$\pm$0.002              & 0.593$\pm$0.003              \\
\hline
\multirow{2}{*}{\textbf{ToM}}    & ×                                   & 0.614$\pm$0.007              & 0.621$\pm$0.010              & 0.611$\pm$0.009              \\
                                 & \checkmark                          & 0.616$\pm$0.008              & 0.619$\pm$0.011              & 0.622$\pm$0.009              \\
\hline
\multirow{2}{*}{\textbf{ToM+EM}} & ×                                   & 0.621$\pm$0.014              & 0.623$\pm$0.011              & 0.620$\pm$0.010              \\
                                 & \checkmark                          & 0.641$\pm$0.012              & 0.639$\pm$0.016              & 0.642$\pm$0.012              \\
\hline
\end{tabular}
\end{adjustbox}
\end{table}

\subsubsection{Effectiveness of Network Modules}

We evaluate the impact of each network module on performance in Table \ref{table_ablation_network_modules}, showing contributions to overall improvement. Fusion alone provides limited gains in model performance but increases stability to some extent. Contrastive learning effectively enhances the model's ability to distinguish between positive and negative samples. Introducing fusion alongside contrastive learning leads to slight further improvements in accuracy and sensitivity, indicating that fusion, when supported by contrastive learning, enhances the model's ability to identify positive samples.

\begin{table}[h]
\scriptsize
\centering
\caption{Ablation Experiments of Network Modules}
\label{table_ablation_network_modules}
\begin{adjustbox}{max width=\textwidth}
\setlength{\tabcolsep}{3pt}
\begin{tabular}{|c|c|c|c|c|}
\hline
\textbf{Fusion} & \textbf{Contrast} & \textbf{Accuracy$\uparrow$} & \textbf{Specificity$\uparrow$} & \textbf{Sensitivity$\uparrow$} \\
\hline
× & × & 0.621$\pm$0.014 & 0.623$\pm$0.011 & 0.620$\pm$0.010 \\
\hline
\checkmark & × & 0.620$\pm$0.007 & 0.621$\pm$0.007 & 0.622$\pm$0.009 \\
\hline
× & \checkmark & 0.641$\pm$0.012 & 0.639$\pm$0.016 & 0.642$\pm$0.012 \\
\hline
\checkmark & \checkmark & \textbf{0.647$\pm$0.008} & \textbf{0.651$\pm$0.009} & \textbf{0.646$\pm$0.009} \\
\hline
\end{tabular}
\end{adjustbox}
\end{table}

\section{Discussion}

Compared to existing methods, BEAM captures shared and unique aspects of ToM and EM, enabling precise and comprehensive modeling of the relationship between signals and empathy levels. The InfoNCE-based contrastive module improves robustness by reducing cross-subject variability and enhancing inter-class separability, while advanced feature fusion integrates complementary information to strengthen feature stability. These innovations make BEAM a valuable tool for empathy research and practical applications.

Table \ref{table_empathy_components} shows that ToM-only inputs outperform EM-only inputs, likely due to ToM's stronger link with understanding intentions and predicting behavior. Individuals with better ToM abilities are more adept at assessing others' needs in social contexts \cite{c18}, whereas EM, though improving perception of others' states, can lead to behavioral withdrawal if over-engaged \cite{c19}. After adding contrastive learning to EM, specificity slightly decreased, likely due to the focus on enhancing the model's ability to distinguish between positive and negative samples.

Although BEAM offers a novel insight into the assessment of children's empathy, it has limitations. First, current label definition focuses on willingness to help, limiting analysis of complex empathy states. This simplification was intentional, considering young children's limited attention span and understanding, requiring clear questions. By focusing on helping behavior, a key expression of empathy \cite{c20} and prosocial behavior, we make empathy more comprehensible for children. Additionally, the dataset size is limited, and individual variations in children's empathy indicate opportunities for optimizing the network structure, particularly in the encoder. Future work should focus on expanding the dataset, refining labeling, and using advanced deep learning techniques to improve model performance and applicability for more accurate empathy assessment and interventions.

\section{Conclusion}
This study proposes BEAM, a model for predicting young children’s empathy levels, addressing challenges such as subjectivity, lack of prediction methods, and individual variability. Using high-resolution EEG and multi-view feature extraction via the LaBraM encoder, feature fusion for improved interpretability, and contrastive learning to enhance robustness. Experiments show that BEAM outperforms SOTA models, with multi-modal inputs further improving performance. These findings highlight the potential of BEAM to advance empathy research and support interventions. Limitations include a small dataset and simplified labels; future work should expand datasets, refine labels, and develop child-specific encoders for more precise empathy modeling.

\addtolength{\textheight}{-12cm}   





\end{document}